\documentclass[letterpaper, 10 pt, hidelinks, conference]{ieeeconf}
\pdfoutput=1
\IEEEoverridecommandlockouts
\usepackage{cite}
\usepackage{amsmath,amssymb,amsfonts}
\usepackage{algorithmic}
\usepackage{graphicx}
\usepackage{textcomp}
\usepackage{hyperref}
\usepackage{comment}
\usepackage{xcolor}
\usepackage{CJKutf8}
\def\BibTeX{{\rm B\kern-.05em{\sc i\kern-.025em b}\kern-.08em
    T\kern-.1667em\lower.7ex\hbox{E}\kern-.125emX}}

\title{\LARGE \bf Gaka-chu: a self-employed autonomous robot artist}
\author{Eduardo Castell{\'{o}} Ferrer$^{1,2}$, Ivan Berman$^{3}$, Aleksandr Kapitonov$^{3}$,\\ Vadim Manaenko$^{3}$, Makar Chernyaev$^{3}$, Pavel Tarasov$^{3}$\\
\small{$^1$MIT Media Lab, Massachusetts Institute of Technology, Cambridge, USA}\\
\small{$^2$MIT Connection Science, Massachusetts Institute of Technology, Cambridge, USA}\\
\small{$^3$M2M Economy, Inc (``Merklebot''), San Francisco, CA, USA}
}

\begin{document}
\maketitle
\urlstyle{same}
\thispagestyle{empty}
\pagestyle{empty}
{\let\thefootnote\relax\footnote{{Corresponding author: Eduardo Castell\'{o} Ferrer, ecstll@mit.edu}}}

\vspace{-0.35cm}
\begin{abstract}
The physical autonomy of robots is well understood both theoretically and practically. By contrast, there is almost no research exploring their potential economic autonomy. In this paper, we present the first economically autonomous robot---a robot able to produce marketable goods while having full control over the use of its generated income. {\it \mbox{Gaka-chu}} (``painter'' in Japanese) is a 6-axis robot arm that creates paintings of Japanese characters from an autoselected keyword. By using a blockchain-based smart contract, {\it \mbox{Gaka-chu}} can autonomously list a painting it made for sale in an online auction. In this transaction, the robot interacts with the human bidders as a peer not as a tool. Using the blockchain-based smart contract, {\it \mbox{Gaka-chu}} can then use its income from selling paintings to replenish its resources by autonomously ordering materials from an online art shop. We built the {\it \mbox{Gaka-chu}} prototype with an Ethereum-based smart contract and ran a 6-month long experiment, during which the robot created and sold four paintings, simultaneously using its income to purchase supplies and repay initial investors. In this work, we present the results of the experiments conducted and discuss the implications of economically autonomous robots.
\end{abstract}

\section{Introduction}
Robots and cyber-physical systems deployed in the real world are reaching increasing levels of autonomy and versatility. They can be programmed to perform tasks with little to no human intervention and can vary significantly in size, functionality, mobility, dexterity and intelligence. In general, physical autonomy (i.e., the ability to observe and act on a physical environment autonomously) is well understood and the state of the art is quite advanced, both theoretically and practically. However, the increasing physical autonomy of devices and systems in the real world opens up new issues beyond the physical environment: for instance, security, accountability, auditability, and other social and ethical issues~\cite{bijani2014review, Akram2017, Awad2018}. In contrast to physical autonomy, very little research has been done on the social and economic autonomy (i.e., the ability to observe and act on a social or economic environment independently) of robots and cyber-physical systems.

The idea of artificial economic autonomy has been explored in philosophical, social, legal, and economic theory. For instance, a line of research considering software-based AI agents has explored the concept of {\it machina economicus}~\cite{parkes2015economic}, or rational AI agents that can reason in economic contexts, either as an ideal synthetic version of the perfectly rational {\it homo economicus}~\cite{levitt2008homo}, or using new incentive mechanisms. Early practical examples of this concept include autonomous software agents for automatic trading~\cite{AutonomousEconomicAgents2020} or AI-driven setting of market prices~\cite{danassis2021achieving}, in which AI agents might perform better economic reasoning than humans. The concept of {\it machina economicus} has also been expanded from artificial agents that mimic humans to the concept of {\it automata economicus}, in which artificial agents achieve new types of economic value creation and build an artificial ``creative economy''~\cite{nobre2018automata}. A few existing works have also looked at economic autonomy specifically in robots. For instance,~\cite{reynolds2008economically} defines a fully economically autonomous robot as one that can use its generated income to cover the costs of its manufacture and maintenance. In the field of management and marketing,~\cite{ivanov2017robot} explores the question of what happens when domestic robots and digital assistants not only make purchases on behalf of their owners, but take on consumer responsibilities such as filtering goods and services for their owner to choose from. The legal implications of economically autonomous robots are also being explored, e.g., robot-based tax systems~\cite{ivanov2017robonomics}, copyright and intellectual property rights of robot creations~\cite{davies2011evolutionary,buning2015eu}, and whether robots need to obey copyright law~\cite{grimmelmann2015copyright}. Although some of these theory developments are more than a decade old, there have been almost no practical advances with real robots, and no prototype of an economically autonomous robot has yet been built.

In this paper, we present the first prototype of an economically autonomous robot, i.e., a robot that can autonomously generate income by producing marketable goods or services and can use the income to autonomously maintain itself through the purchase of resources. Our economically autonomous robot {\it \mbox{Gaka-chu}} (``painter'' in Japanese) is a self-employed robot artist that autonomously makes physical paintings, sells them in online auctions, and uses the income it generates to purchase art supplies from an online shop and repay start-up loans from initial investors. We built {\it \mbox{Gaka-chu}} using blockchain technology, through rules encoded as Smart Contracts (SCs)---computer code embedded in the blockchain that directly controls the transfer of digital assets between parties~\cite{buterin2014next}. The control logic for the robot resides in the SC while the actuation takes place in the physical world. 

\subsection{Blockchain-based robotics}
\label{ssec:blockchain-robots}
% Description of ADEPT and Plantoid
The first proof-of-concept robotics system that used blockchain technology to make financial transactions was the ADEPT~\cite{veena2015empowering} protocol introduced by Samsung and IBM. The project focuses on increasing the autonomy of devices or machines that operate in a decentralized manner within (industrial) IoT. For their proof-of-concept, they used a washing machine (W9000) that could autonomously order detergent every time it ran out. The ADEPT project also led to a pilot of a Blockchain-of-Devices (BoD), where devices work together autonomously and make decisions about tasks or orders~\cite{Higgins2015}. Along these lines, the {\it Plantoid (2015)} art project by Okhaos\footnote{News articles on the {\it Plantoid (2015)} art project by Okhaos: in {\it Furtherfield}, ``Plantoid: The Blockchain-Based Art That Makes Itself'' by Robert Myers, available at \url{https://www.furtherfield.org/tag/autopoesis}, and in {\it Coindesk}, ``This Robot Plant Needs You and Bitcoin to Reproduce'' by Grace Caffyn, available at \url{https://www.coindesk.com/this-robot-plant-needs-you-and-bitcoin-to-reproduce}.} proposed a metallic robotic sculpture designed to look and move like a plant and be displayed in a public space. If humans enjoy the sculpture and make a small donation to it, the sculpture dances with plant-like movements, music, and lights. Contributions are made via the Bitcoin blockchain. Once the sculpture has collected a sufficient quantity of Bitcoins in its crypto wallet, it asks humans to help it by reproducing it (creating a new sculpture) and placing the ``offspring'' in a new location. 

% Difference between this project and previous approaches
However, in the first example (ADEPT), the washing machine did not have control over its income nor the values used to complete tasks or make the orders, and in the second example (Plantoid), the robotic sculpture did not produce external goods or services nor did it make consumer choices and purchases autonomously. Moreover, its donation-based model might also lead to long wait times and deadlocks. By contrast, an economically autonomous robot must have control over the action of entering the market to offer goods or services and over the consumer purchases it makes using the income that it generates. Therefore, although these two projects are important precedents in devices making transactions using blockchain-based technology, they are not economically autonomous robots.

Beyond economic autonomy, recent research has demonstrated many security and robustness benefits of combining autonomous robots with blockchain-based technology~\cite{Strobel2018, Strobel2020, Ferrer2021}, using both SCs and Merkle trees. In these research works, the robots are used as nodes in a network and their interactions are encapsulated in cryptographic transactions. Blockchain technology can give data confidentiality and entity validation to robots~\cite{Ferrer2018RoboChain, Malsa2021}, making them suitable for applications in which privacy and security are a concern~\cite{Castello2021, Rainer2022}.

In summary, previous literature has overlooked a prototype of an economically autonomous robot that not only directly manages its income generation and the resources it needs to maintain its creation of economic value (e.g., recharging batteries, paying for supplies), but also adapts to market needs. Historically, robots have participated in labor roles (e.g., in factories, assembly lines), but with new decentralized financial tools such as blockchain-based SCs, robots can now also take part in other aspects of our economic environment, redefining their role as not mere tools but potential peers. 

% Fig 1 (overview)
\begin{figure}[tbh]
\vspace{0.1cm}
\centering
\includegraphics[width=\columnwidth]{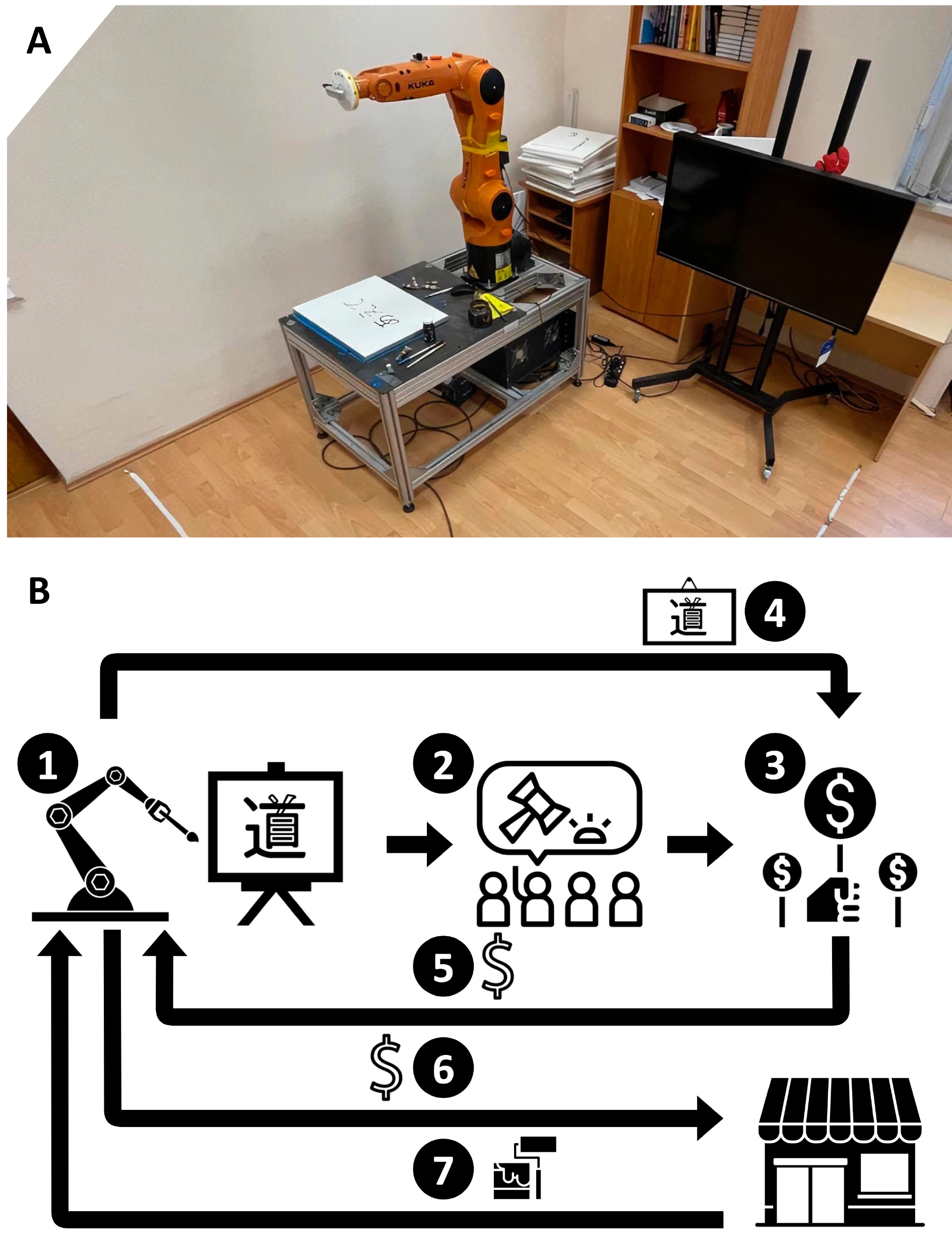}
\caption{A) {\it \mbox{Gaka-chu}} (``painter'' in Japanese), a 6-axis robot arm (KUKA KR6 R900) programmed using an Ethereum-based SC while painting a canvas. {\it \mbox{Gaka-chu}}'s work area dimensions are $2.53\times2.57$ meters and are specially designed to maximize safety. B) Typical workflow for a painting job: (1) The robot uses its sensing, computation, and actuation capabilities to paint a canvas. (2) The robot puts the painting up for auction, participants can place their bids in the auction website. (3) A winner is selected when the painting process is finished and the final bid is deposited (by using ETH: the cryptocurrency of the Ethereum network). (4) Ownership of the painting is transferred to the winner of the auction. (5) The payment is transferred to the robot's account. (6)-(7) The robot can order supplies from an art shop to maintain its painting activities.}
\label{fig:InitialDiagram}
\end{figure}

\begin{figure*}[ht!]
    \vspace{0.1cm}
    \centering
    \includegraphics[width=\linewidth]{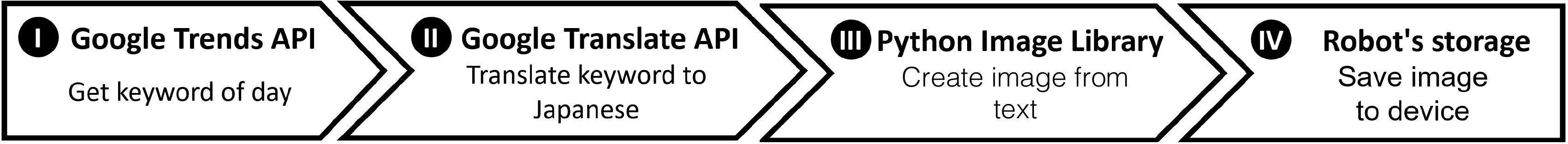}
    \caption{Workflow for topic selection: {\it \mbox{Gaka-chu}} first (I) selects the English keyword with the largest number of searches that day in Google Trends, (II) translates the keyword into Japanese characters, (III) converts the resulting text into an image, and (IV) uses the image as an input for the physical painting process.}
    \label{fig:SelectingTopics}
    \vspace{-5mm}
\end{figure*}

\subsection{{\it \mbox{Gaka-chu}}: a self-employed autonomous robot artist}
\label{ssec:gakachu}
% Main problems tackled and overview of the results the reader is going to find
In this paper, we present an economically autonomous robot named {\it \mbox{Gaka-chu}}. {\it \mbox{Gaka-chu}} paints canvases, uses blockchain-based technology to sell the paintings it makes in online auctions, generates and collects income, and uses its income to purchase the material resources needed to maintain its activity, all with minimal human intervention (Fig.~\ref{fig:InitialDiagram}). Finally, {\it \mbox{Gaka-chu}} is able to pay back initial investors. The three main challenges of an economically autonomous robot are that: (1) the robot should generate income to maintain itself, (2) the robot should have a mechanism that adapts to market changes, and (3) when interacting with third-party agents such as online shops or human bidders, the robot should do so as a peer, not as a tool. In this paper, {\it \mbox{Gaka-chu}} meets these three main challenges, demonstrated in a 6-month experiment. Our findings show that {\it \mbox{Gaka-chu}} can reach economical autonomy: fulfil a job, get rewarded for it, and invest the benefits in its own sustainability.

% Explanation about what the reader is going to see in the following sections
The remainder of this paper is structured as follows. Section~\ref{sec:implementation} describes (\ref{ssec:topic}) how the robot selects a topic for painting, (\ref{ssec:drawing}) the sensing, planning, and actuation when painting a canvas, (\ref{ssec:selling}) how the online auctions are organized and financial transactions made, and (\ref{ssec:supplies}) how the robot interacts with the autonomous online shop to provide the robot with the necessary painting consumables. Section~\ref{sec:results} presents the results of a 6-month experiment including a start-up ``loan'' from human investors, making and auctioning four paintings, and fulfilling financial transactions with webshops and human peers (suppliers, and customers). Section~\ref{sec:discussion} discusses the results, and the implications of economically autonomous robots, and proposes future directions. Finally, Section~\ref{sec:conclusions} provides our conclusions.

\section{Methods}
\label{sec:implementation}
As any artist painter, {\it \mbox{Gaka-chu}} has four main challenges: (1) how to select a topic for the painting, (2) how to actually paint it, (3) how to sell the painting to obtain economic resources, and finally, (4) how to use the generated income to purchase the necessary materials to continue with its activity. All involved software described in this section is available at our public GitHub repository\footnote{Github repository: \url{https://github.com/Multi-Agent-io/gaka-chu.online}}.

\subsection{Selecting a topic for a painting}
\label{ssec:topic}
Each painting by {\it \mbox{Gaka-chu}} consists of a set of Japanese characters which together form a keyword. {\it \mbox{Gaka-chu}} adapts to the market by choosing the keywords that are popular in Google Trends\footnote{Google Trends: \url{https://trends.google.com}}. The topic selection process (Fig.~\ref{fig:SelectingTopics}) proceeds as follows: (I) {\it \mbox{Gaka-chu}} requests the keyword with the maximum number of searches for the current day from the Google Trends API. (II) {\it \mbox{Gaka-chu}} translates the selected keyword to Japanese using the Google Translate\footnote{Google Translate: \url{https://translate.google.com}} API. (III) {\it \mbox{Gaka-chu}} converts the text-based Japanese characters into an image of black strokes (forming the Japanese words) in the center of a white background, using the Python Image Library (PIL)~\cite{umesh2012image}, and (IV) saves the image for further processing. In this approach, {\it \mbox{Gaka-chu}} chooses topics that are generally popular and potentially profitable for sale, and still the paintings are not associated with any previous copyright claims.

\subsection{Painting process}
\label{ssec:drawing}
\begin{figure}[tbh]
    \centering
    \includegraphics[width=\columnwidth]{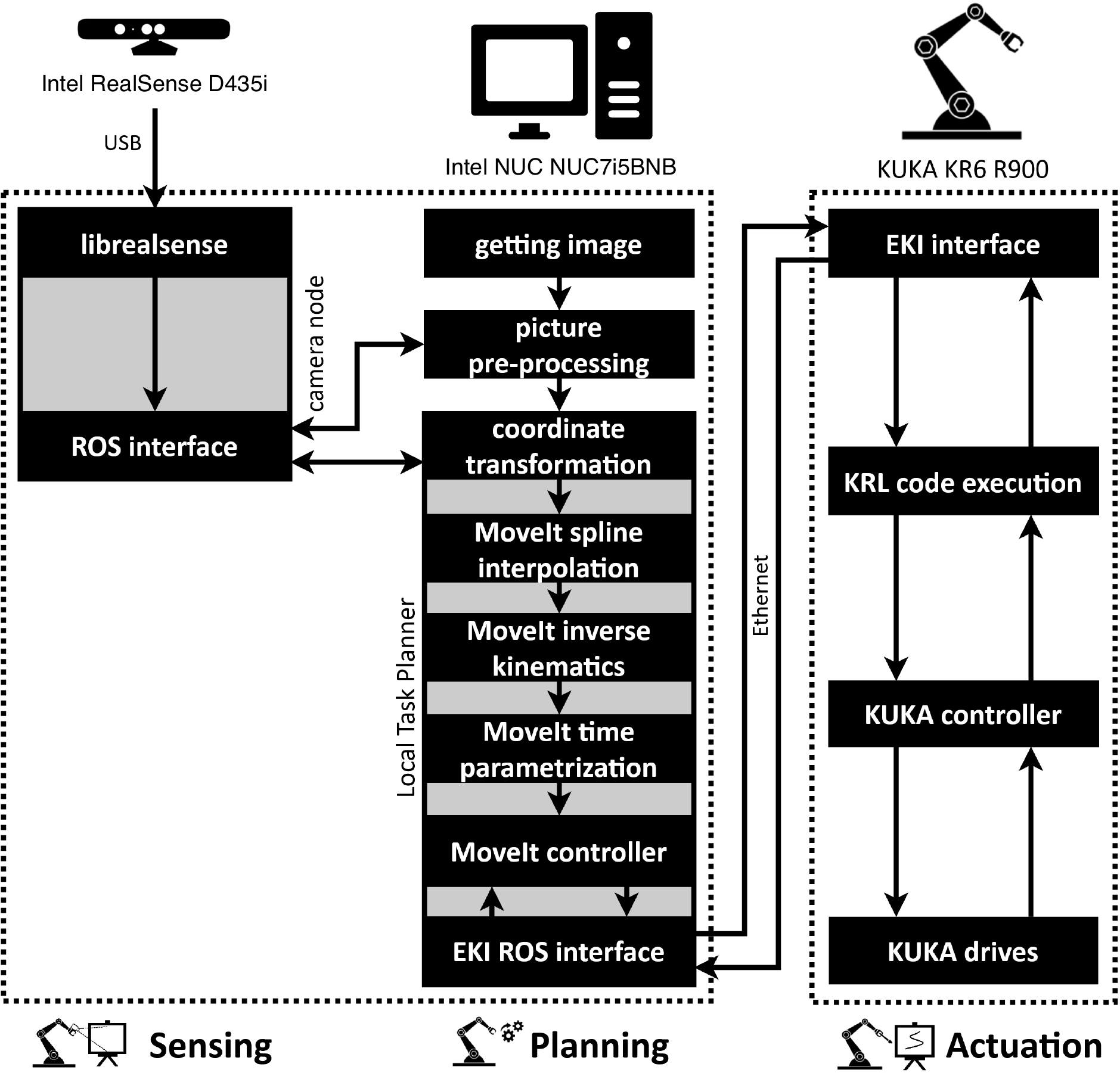}
    \caption{Functional scheme and workflow for the {\it \mbox{Gaka-chu}} painting process. In summary, the sensing part gets the information about the canvas (e.g., position, orientation) from a depth camera installed at the end-effector. This information is sent to the planning part, which runs in an Intel NUC single-board computer and calculates all trajectories for the {\it \mbox{Gaka-chu}} joints. Finally, in the actuation part, calculated trajectories are sent to the internal motor controllers, which execute the movements.}
    \label{fig:FBProcess}
\end{figure}

After an image of the selected keyword is created and stored, {\it \mbox{Gaka-chu}} starts the physical painting process. The three main components---sensing, planning, and actuation---are depicted in Fig.~\ref{fig:FBProcess}.

\subsubsection{Sensing}
\label{sssec:sensing}

\begin{figure}[tbh]
    \vspace{0.2cm}
    \centering
    \includegraphics[width=\columnwidth]{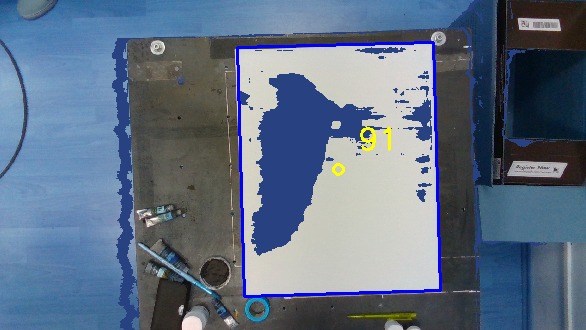}
    \caption{An example of canvas detection using the Intel RealSense D435i depth camera. The image shows the angle of rotation of the canvas relative to {\it \mbox{Gaka-chu}}'s horizontal axis.}
    \label{fig:CameraOutput}
\end{figure}

In the sensing pipeline (see the left block of Fig.~\ref{fig:FBProcess}), a depth camera (Intel RealSense D435i) is mounted on {\it \mbox{Gaka-chu}}'s end-effector and connected to a single-board computer (Intel NUC7i5BNB) via USB.
The position and orientation of the canvas is detected from the point cloud information from the depth camera (Fig.~\ref{fig:CameraOutput}). This process is handled by a ROS \cite{Quigley2009ROS:System} camera node, working in client-service mode and based on the realsense library\footnote{\url{https://dev.intelrealsense.com}}. The ROS camera node also publishes the dimensions and center point of the canvas which are necessary for the correct coordinate transformation from the canvas to the camera frames. It then provides this information upon request to the planning component of the system through its ROS interface.

\subsubsection{Planning}
\label{sssec:planning}

\begin{figure}[tbh]
    \centering
    \includegraphics[width=\columnwidth]{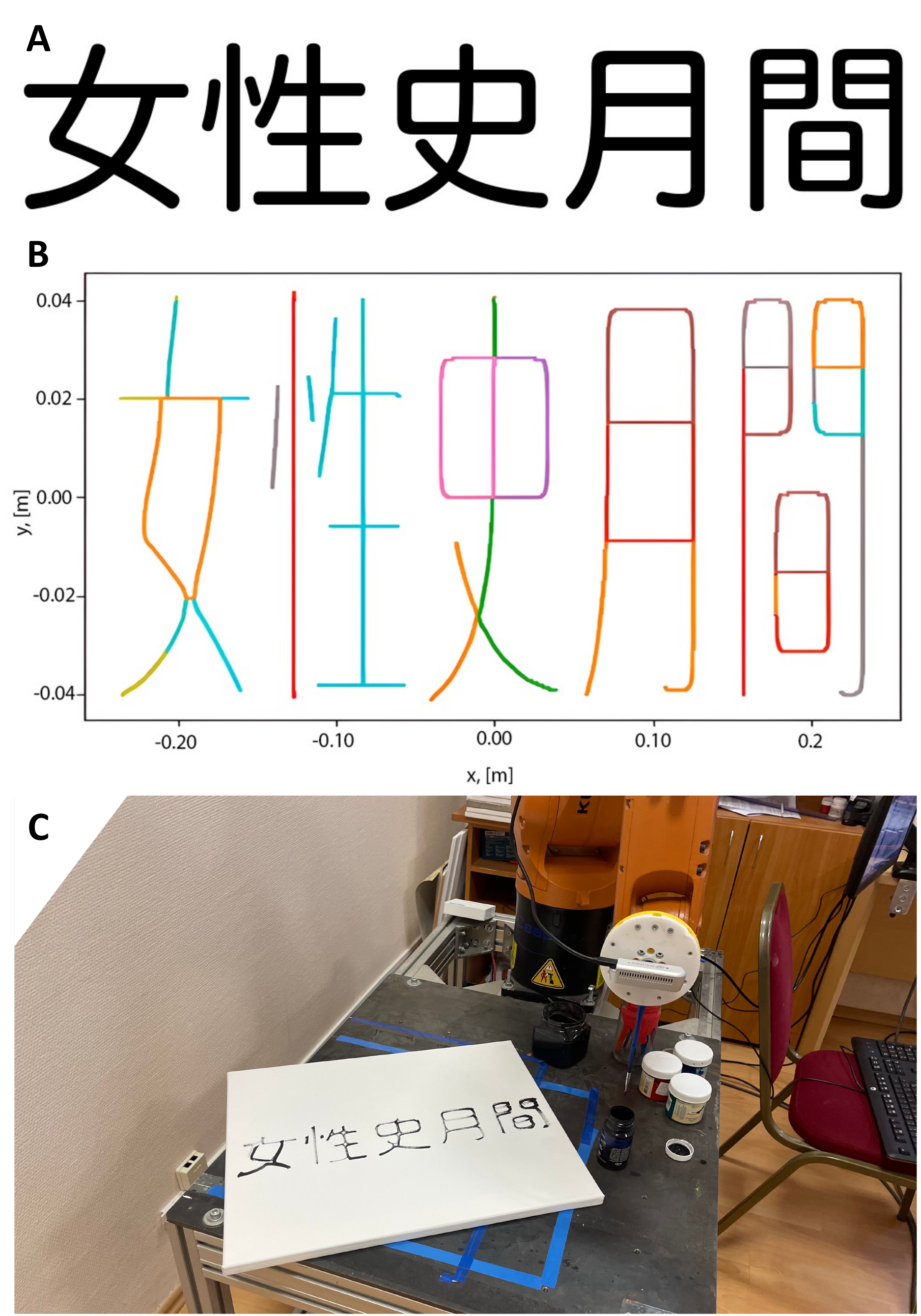}
    \caption{An example of the skeletonization and painting process. (A) 5 characters form the Japanese term \begin{CJK}{UTF8}{min}女性史月間\end{CJK} obtained after translating the Google Trends keyword ``Women's History Month.'' (B) The same image after skeletonization of the 2D shapes into 1D curves and conversion from pixels to meters. Different colors represent different strokes in the painting process. (C) An image of the finished painting. }
    \label{fig:skeletonization}
\end{figure}

In the planning pipeline (see the center block of Fig.~\ref{fig:FBProcess}), {\it \mbox{Gaka-chu}} calculates the end-effector trajectories needed to paint the image generated in Sec.~\ref{ssec:topic}. First, {\it \mbox{Gaka-chu}} pre-processes the stored image by converting the binary 2D shapes of the Japanese characters into 1D curves, i.e., ``skeletonization''~\cite{saha2017skeletonization} (Fig.~\ref{fig:skeletonization} A and B), using the OpenCV library~\cite{opencv_library}. The 1D curves are represented as pixels in a 2D coordinate frame. The pixel coordinates are passed to the Local Task Planner (LTP), which translates the pixels of the skeletonized image from their starting image-based coordinate frame to {\it \mbox{Gaka-chu}}'s canvas coordinate frame, with the $z$ axis taking into account the 3D position of the canvas relative to {\it \mbox{Gaka-chu}}. Then, the LTP coverts the coordinates from pixels to meters. At this point, {\it \mbox{Gaka-chu}} has all the necessary coordinates to paint the strokes that will form the desired painting. 

For {\it \mbox{Gaka-chu}}'s motion planning, we use the MoveIt Framework~\cite{gorner2019moveit}, with a customized MoveIt module for inverse kinematics. Our custom module extends the standard module by taking into account that the manipulator in different positions has different kinetic energy (i.e., the moment of inertia changes). This would eventually lead to blockage of the movement by the KUKA internal controller. First, the planner interpolates paths to build splines relative to the end-effector coordinate system. Our MoveIt module solves the inverse kinematics and obtains paths for each link of the robot. The motion planner then conducts time parameterization, taking into account the maximum speeds and accelerations for each link of the robot, and finally prepares the calculated trajectories to be sent from ROS to the Ethernet KRL Interface (EKI)\footnote{EKI is a TCP/IP protocol which allows the exchange of XML and binary data between the manipulator and the external system.}. 

\subsubsection{Actuation}
\label{sssec:actuation}
In the actuation pipeline (see the right block of Fig. \ref{fig:FBProcess}), {\it \mbox{Gaka-chu}}'s internal computer runs executable code written in the KUKA Robot Language (KRL)---a programming language similar to Pascal. The KRL program receives the trajectory information calculated in the planning pipeline through the EKI interface and organizes the transfer of the necessary rotation angles, speeds, and accelerations to the KUKA drives. If the controller detects any problems during the movement (e.g., physical impossibility of movement), the manipulator stops and the corresponding error messages are sent back.

The technique {\it \mbox{Gaka-chu}} uses to paint the canvas is based on the existing literature on drawing with strokes~\cite{lindemeier2015hardware,scalera2018busker}. The paint cup is installed at a known position, to one side of the manipulator. The painting algorithm is organized as follows: (1) the robot moves to the position above the paint cup, (2) it dips the brush into the paint and withdraws it, (3) it moves to the starting point of the trajectory with a slight offset along the $z$ axis, (4) it descends to the starting point, (5) it paints one segment of the trajectory, and (6) it raises the brush and repeats the process until the whole painting is complete (Fig.~\ref{fig:skeletonization} C). After testing several types of paints and brushes, we realized that nylon and bristly brushes as well as acrylic paint show the best finishing quality.

\begin{figure}[tbh]
    \vspace{0.1cm}
    \centering
    \includegraphics[width=\columnwidth]{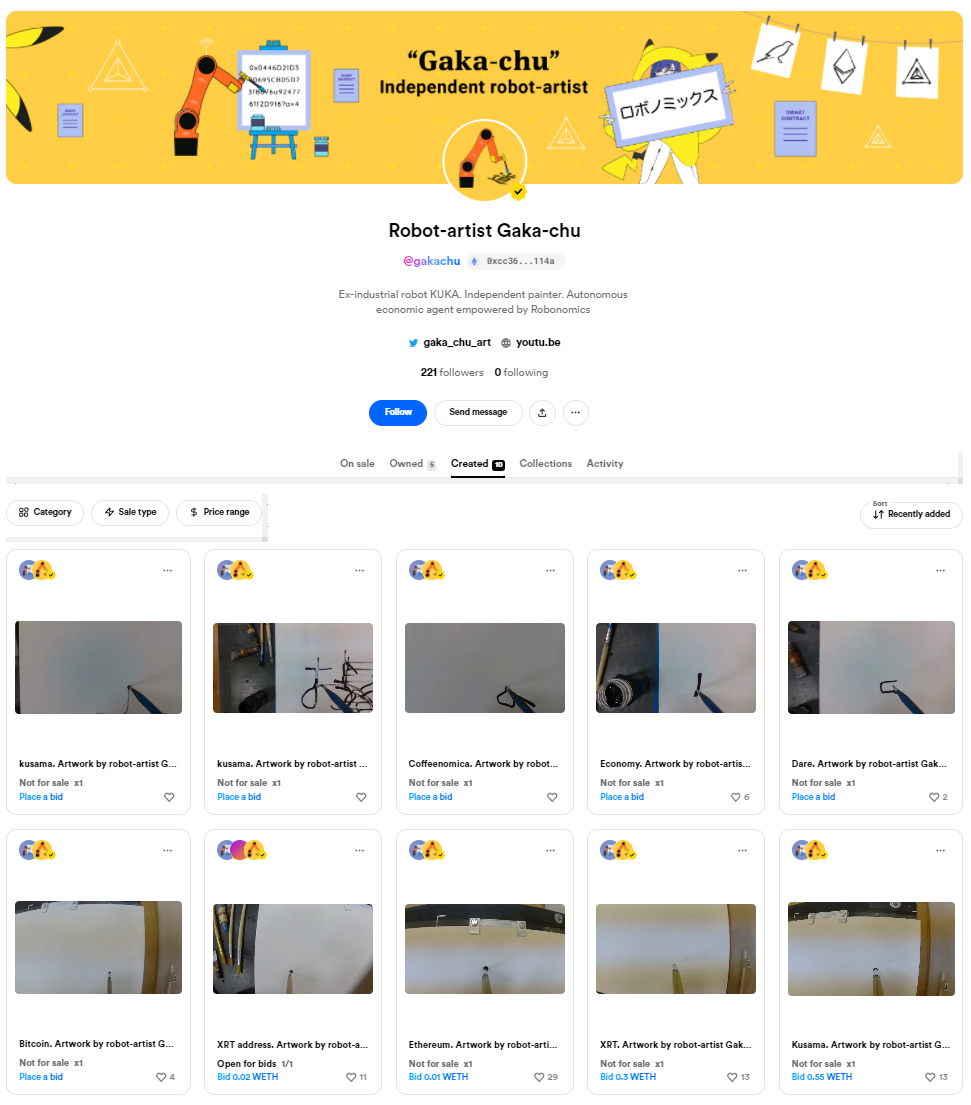}
    \caption{Screenshot of the auction site where {\it \mbox{Gaka-chu}} advertises the completed paintings. Potential customers bid with their own blockchain-based assets until the auction closes and the highest bidder wins the auction lot. Four completed auctions, illustrated in the balance graph (Fig.~\ref{fig:dare-funds-stats}), are located at the very end of the list.}
    \label{fig:auctions}
\end{figure}

\subsection{Generating income: selling paintings}
\label{ssec:selling}

After the painting has been completed, {\it \mbox{Gaka-chu}} puts it up for sale in an online auction (Fig.~\ref{fig:auctions}). The online auctions are hosted on an external service, where {\it \mbox{Gaka-chu}} maintains its own page\footnote{Rarible auction page: \url{https://rarible.com/gakachu}}. The logic of the auction is encoded in an Ethereum-based SC\footnote{\url{https://docs.rarible.org/ethereum/smart-contracts/smart-contracts/}} and is uploaded in the Ethereum Mainnet blockchain. 

When a painting is completed, {\it \mbox{Gaka-chu}} advertises its availability by creating a dedicated tab for it within its own page and uploading a video of the entire painting process. Human bidders are then able to place bids in a given time limit in ETH (i.e., the cryptocurrency of the Ethereum blockchain) using their own crypto wallets. Once the auction is closed, the ownership of the painting is assigned\footnote{Digital ownership is assigned using a non-fungible token (NFT). More information is available at \url{www.gaka-chu.online}.} to the Ethereum address of the highest bidder, and the bid amount is sent to {\it \mbox{Gaka-chu}}'s Ethereum address. The process of arranging shipping and delivery of the physical painting to the mailing address of the highest bidder is of course arranged off-chain, with the manual help of a human assistant.  

\subsection{Using income: purchasing consumables and paying back investors}
\label{ssec:supplies}
{\it \mbox{Gaka-chu}} can use its generated income to order the art materials needed for future painting jobs (e.g., canvases, paints, brushes). {\it \mbox{Gaka-chu}} autonomously places orders at an online art shop that accepts payments in ETH by using a web-app we developed\footnote{Art shop interface: \url{https://old.dapp.robonomics.network/\#/art-shop}}. {\it \mbox{Gaka-chu}} decrements its canvas stock counter each time it completes a painting, and decides to purchase a set of supplies (canvases, paints, and brushes) when the counter reaches one. Using this software, {\it \mbox{Gaka-chu}} generates a message containing the composition of the order, the Ethereum address, and the amount of ETH tokens to be paid, and sends the message to the art shop through its server API. If the parameters for the purchase are acceptable, the art shop agrees to execute it. All steps in this two-way communication are conducted through an Ethereum-based SC, obliging the art shop to fulfill its liabilities to provide consumables. The SC ends with the sending of a final message about a successful purchase and the sending of ETH tokens to the shop's Ethereum wallet. When supplies are received by post, a human assistant places the canvas, paint, and brush in their positions. Finally, after successfully selling a batch of paintings, {\it \mbox{Gaka-chu}} pays back the initial investments that were used to start its activity (e.g., pay for the auction site platform and sign-up fees).

\section{Results}
\label{sec:results}
To demonstrate the feasibility of an economically autonomous robot, we ran a 6-month experiment in which {\it \mbox{Gaka-chu}} received a starting ``loan'' from early human investors, painted canvases and sold them at auction to human bidders, used the generated income to maintain its economic activity by purchasing supplies, and paid back its human investors. During the experiment, {\it \mbox{Gaka-chu}} painted four canvases which were put up and sold at online auctions. 

\begin{figure*}[htb]
    \centering
    \includegraphics[width=1.98\columnwidth]{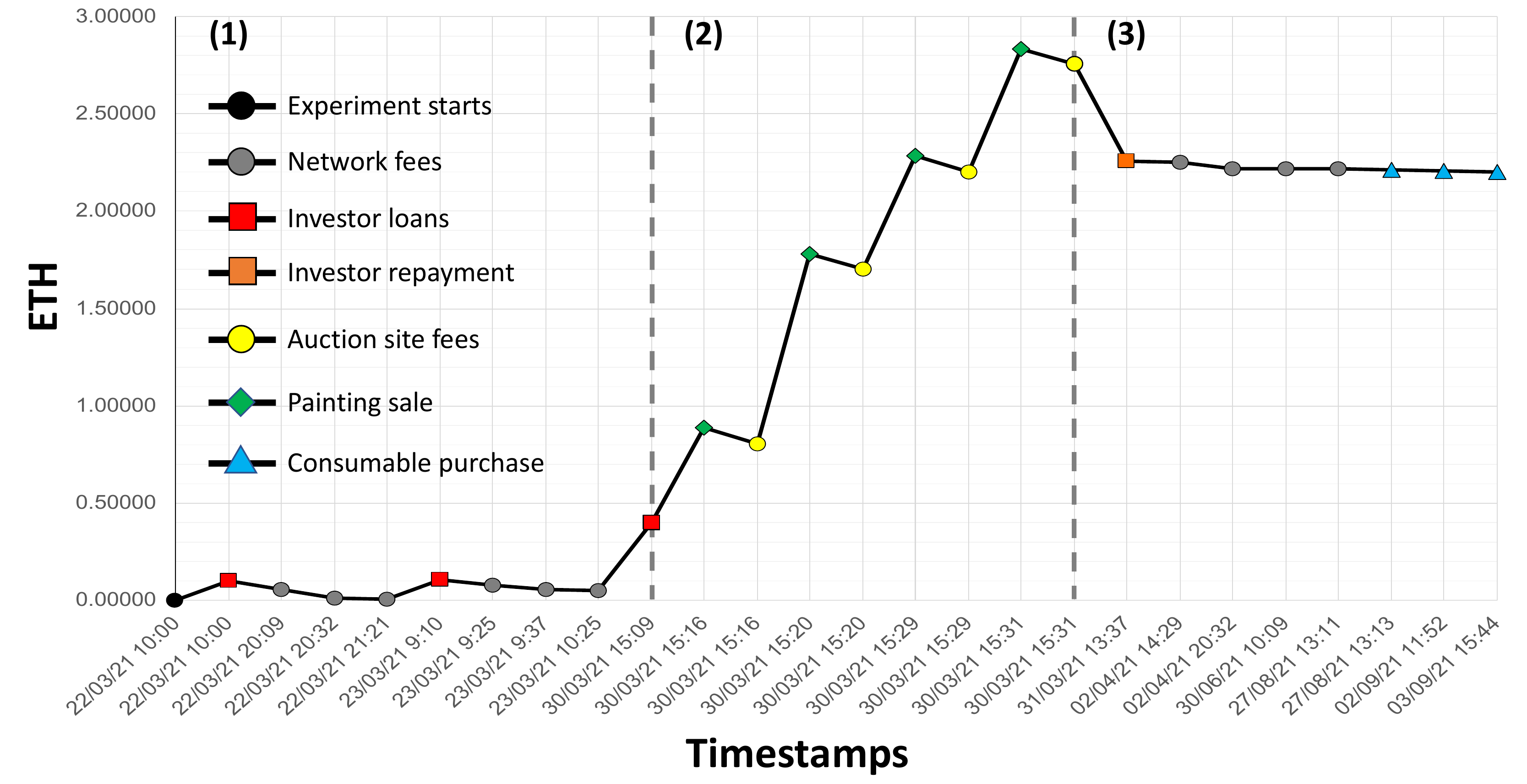}
    \caption{Balance on {\it \mbox{Gaka-chu}}'s Ethereum wallet from March 22\textsuperscript{nd} 2021 to September 3\textsuperscript{rd} 2021. The graph is organized as a progressive line chart where the different stages of the experiment are depicted: (1) funding, (2) auctioning and (3) investment repayment and supplies acquisition. In the chart, red squares represent transactions from initial investors to fund the system. Green diamonds represent the sale of a painting. Blue triangles represent the purchase of consumables and grey dots represent the payment of Ethereum network service fees. Yellow dots represent platform service fees for the auction site. Finally, orange squares represent loan repayments to initial investors.}
\label{fig:dare-funds-stats}
\vspace{-5mm}
\end{figure*}

Figure~\ref{fig:dare-funds-stats} shows the balance changes on {\it \mbox{Gaka-chu}}'s Ethereum wallet\footnote{{\it \mbox{Gaka-chu}}'s Ethereum wallet address:\\ {\fontsize{7.75}{10}\selectfont \url{https://etherscan.io/address/0xcc3672c869c923b90f2c1bfba2c7801e3924114a}}} during a 6-month period of time (March to September 2021). Figure~\ref{fig:dare-funds-stats} is organized as a line chart where the different stages of the experiment are depicted: (1) funding, (2) auctioning and (3) investment repayment and supplies acquisition. Initially, in order to setup an account on the auction site, {\it \mbox{Gaka-chu}} requires to send several transactions and pay network fees (grey dots). For that purpose, {\it \mbox{Gaka-chu}} received an initial ``loan'' from human investors (red squares transactions). {\it \mbox{Gaka-chu}} then sold four completed paintings at different auctions, transferred the ownership of the paintings to the highest bidders, and in return received the corresponding bid amount in ETH (green diamonds). {\it \mbox{Gaka-chu}} was then able to pay back its human investors (orange square), purchase more art supplies (blue triangles) and pay the auction site fees (yellow dots). The videos of all {\it \mbox{Gaka-chu}}'s completed paintings and results of its four closed auctions can be retrieved from the auction host website.

\section{Discussion}
\label{sec:discussion}
Robots of all types, from autonomous vehicles to 3D printing devices, are revolutionizing a wide variety of industries: mobility, manufacturing, and logistics. The emergence of robotics is commonly acknowledged as one of the main disruptive changes that will have substantial socioeconomic impact in upcoming decades~\cite{Yang2018a}. As countries and companies adjust to increasing social and economic shifts, interest in robotics will continue to increase---automation is key to improving productivity. In the one hand, financial software-based agents (e.g., Paypal, High Frequency Trading, eCommerce) have had a huge impact on productivity, but they do not impact the physical realm directly. On the other hand, modern robots and cyber-physical systems interact with the physical world in powerful ways (i.e., sense, plan, and actuate), but they lack the technical ability to directly participate in financial transactions due to insufficient agency. 

However, blockchain-based technologies such as SCs allow robots to become online, digital economic actors that can financially operate with minimal human intervention. The combination of both fields opens the door towards financial autonomous robots that can redefine their role as not mere tools but potential peers. This opens a debate not only about the concept of property and ownership under increasing robot autonomy~\cite{michalski_robotsue} but also whether these systems can become a future cornerstone of societal economic activity (e.g., a Universal Basic Income~\cite{straubhaar2017economics} through taxation of robot's economical activities). 

For future work, we would like to focus on improving the selection of the keyword to paint. We consider that after a series of sales, the topic selection strategy could be adjusted based on historical data. In addition, we would like {\it \mbox{Gaka-chu}} to move beyond calligraphy paintings and let it explore other artistic styles (e.g., abstract art, caricature portraits, and pointillism paintings) and modes of production. Finally, financially autonomous robots also need to be able to make purchases to maintain their continued viability. For instance, future robotics systems using this technology might need to be able to pay for electricity (e.g., to a street charger), spare or replacement parts (e.g., to Mouser), training data (e.g., to MTurk), or skilled human resources (e.g., to TaskRabbit) with their blockchain-based financial assets. Developing new payment gateways and interfaces to these services is a promising way forward. 

\section{Conclusions}
\label{sec:conclusions}
In the robotics research field physical autonomy (i.e., the ability to observe and act on a physical environments) is well understood and the state of the art is advanced. However, very little research has been done on the economic autonomy of robots. In this work, we have presented and demonstrated the first economically autonomous robot, using blockchain-based smart contracts to build {\it \mbox{Gaka-chu}}, a self-employed autonomous robot artist. In the proposed system, the control logic for the robot resides in the smart contract uploaded to the Ethereum blockchain while the actuation takes place in the physical world. In this research, {\it \mbox{Gaka-chu}} makes physical paintings, sells them to humans in online auctions, and uses the income it generates to purchase art supplies from an online shop, and pay back initial investors. Our findings show that after a 6-month long experiment including a start-up ``loan'', making and auctioning four paintings, and making financial transactions with webshops and human peers (suppliers, and customers), {\it \mbox{Gaka-chu}} can reach economic autonomy: fulfil a job, get rewarded for it, and invest the benefits in its own sustainability.

\bibliography{references.bib}
\bibliographystyle{IEEEtran}
\end{document}